\documentclass[fleqn,10pt]{wlscirep}
\usepackage[utf8]{inputenc}
\usepackage[T1]{fontenc}

\usepackage{placeins}
\usepackage{hyperref}
\usepackage{verbatim}

\usepackage{xr}
\hypersetup{
pdftitle={A Hybrid DL-Model for ENSO},
pdfsubject={preprint},
pdfauthor={Jakob Schloer, Matthew Newman, Jannik Thuemmel, Antonietta Capotondi, Bedartha Goswami},
pdfkeywords={El Ni\~{n}o Southern Oscillation, Deep Learning, Linear Inverse Models},
}
\begin{document}

%TC:ignore
%\detailtexcount{hybrid_lim}
%TC:endignore
\newpage

%TC:ignore
\title{A Hybrid Deep-Learning Model for El Niño Southern Oscillation in the Low-Data Regime}

\author[1,2,*]{Jakob Schloer}
\author[3]{Matthew Newman}
\author[1]{Jannik Thuemmel}
\author[3,4]{Antonietta Capotondi}
\author[1,5]{Bedartha Goswami}

\affil[1]{Machine Learning in Climate Science, University of Tübingen, Germany}
\affil[2]{European Centre for Medium Range Weather Forecasts (ECMWF), Reading, UK}
\affil[3]{NOAA Physical Sciences Laboratory, Boulder, CO, USA}
\affil[4]{Cooperative Institute for Research in Environmental Sciences, University of Colorado, Boulder, CO, USA}
\affil[5]{Data Science Department, Indian Institute of Science Education and Research, Pune}
\affil[*]{jakob.schloer@uni-tuebingen.de}

\begin{abstract}
    While deep-learning models have demonstrated skillful El Niño Southern Oscillation (ENSO) forecasts up to one year in advance, they are predominantly trained on climate model simulations that provide thousands of years of training data at the expense of introducing climate model biases. Simpler Linear Inverse Models (LIMs) trained on the much shorter observational record also make skillful ENSO predictions but do not capture predictable nonlinear processes. This motivates a hybrid approach, combining the LIM’s modest data needs with a deep-learning non-Markovian correction of the LIM. For O(100 yr) datasets, our resulting Hybrid model is more skillful than the LIM while also exceeding the skill of a full deep-learning model. Additionally, while the most predictable ENSO events are still identified in advance by the LIM, they are better predicted by the Hybrid model, especially in the western tropical Pacific for leads beyond about 9 months, by capturing the subsequent asymmetric (warm versus cold phases) evolution of ENSO.
\end{abstract}

\flushbottom
\maketitle
%TC:enignore
\thispagestyle{empty}

%%%%%%%%%%%%%%%%%%%%%%%%%%%%%%%%%%%%%%%%%%%%%%%%%%%%%%%%%%%%%%%%%%%%%
% MAIN BODY OF PAPER
%%%%%%%%%%%%%%%%%%%%%%%%%%%%%%%%%%%%%%%%%%%%%%%%%%%%%%%%%%%%%%%%%%%%%

\section{Introduction}

% Background
In the past few years, deep learning has revolutionized weather forecasting with models such as GraphCast \citep{lam2023}, Pangu \citep{bi2023}, and others \citep{pathak2022, lang2024} now outperforming traditional state-of-the-art numerical weather prediction models \citep{ben-bouallegue2023}. These advancements are largely due to the availability of millions of training data samples from reanalysis products on hourly resolution, which allows the training of large neural networks with negligible generalization error. While these models demonstrate exceptional medium-range forecasting skill, it remains unclear if these capabilities extend to seasonal or annual predictions. Unlike medium-range forecasting, which is mainly dependent on initial conditions, long-range predictions are primarily determined by boundary forcing factors, particularly from the ocean.

% Related work
A prominent example of a target for seasonal to annual forecasting is El Niño-Southern Oscillation (ENSO). Characterized by tropical Pacific sea surface temperature anomalies (SSTA), the ENSO strongly influences global weather patterns and is therefore a primary source of seasonal to annual predictability. Its events are characterized by anomalously warm (cold) tropical SSTA, which exhibit a rich diversity in their spatial structure, temporal evolution \citep{capotondi2015,timmermann2018}, and impact on extreme weather conditions worldwide \citep{taschetto2020,strnad2022}. Thus, early forecasts of the likelihood of an ENSO event and also of its expected spatial structure are of great value for sectoral applications worldwide \citep{callahan2023}. 

%% DL for ENSO
While deep learning models have demonstrated their capacity for producing skillful ENSO forecasts \citep{ham2019, petersik2020, cachay2021, zhou2023}, training these models directly from SSTA data is hampered by the short observational record. Global SST records from satellites have been available for the past 40 years, with reconstructions based on point observations extending 100 years prior. In fact, \citet{wittenberg2009} suggested that O(500) years of data might be required to entirely capture the diversity of interannual ENSO events. Instead, deep learning models have been trained on lengthy simulations made by global circulation models (GCMs) \citep{ham2019,zhou2023}, such as those available from the CMIP5 or CMIP6 ensembles. While potentially providing hundreds to thousands of years of data, GCM simulations possess inherent biases likely due to their insufficient spatial resolution and to inadequate parametrizations of unresolved processes. In the Tropical Pacific, such biases typically include mean errors in the equatorial upwelling region in the eastern Pacific and variability errors in the strength and location of ENSO anomalies \citep{capotondi2013,chen2021,beverley2023}, all of which may then be captured by the neural network. Transfer learning, as for instance used by \citet{ham2019}, presents a potential workaround for bridging the gap between erroneous simulation data and limited observations. However, Zhou and Zhang \citep{zhou2023} found no significant improvement in model performance when fine-tuning their model with reanalysis data, indicating the need for more research in this domain.

%TC:ignore
\begin{figure}[!t]
    \centering
    \includegraphics[width=1.0\textwidth]{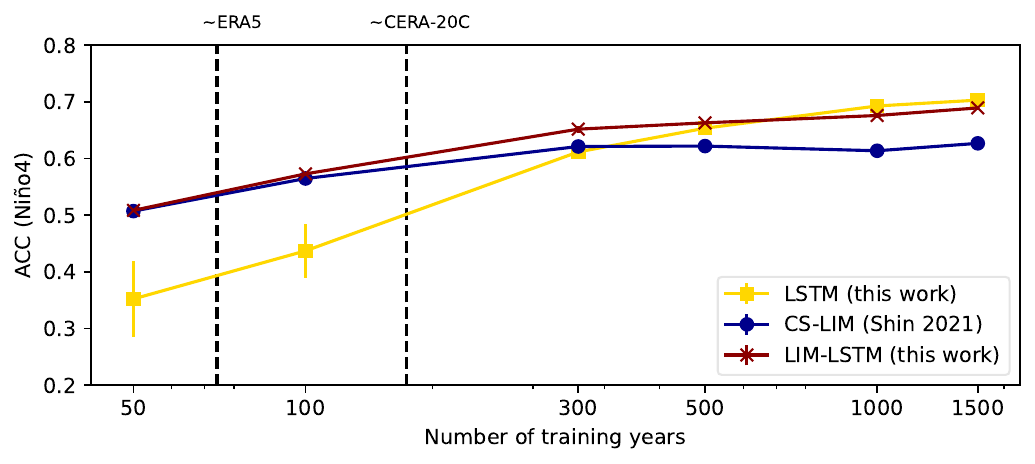}
    \caption{\textbf{Variations in forecast skill over training data length.} 
    The forecast skill of the CS-LIM, LIM-LSTM hybrid model, and LSTM changes with number of years in the training data. Models are trained on random subsets, ranging from 50 - 1500 years, of the training set. The anomaly correlation coefficient (ACC) of the Niño4-index is computed for a forecast lead time of 12-months over the 200-year test period (see Methods Sec.~\ref{sec:data}).}
    \label{fig:training_data}
\end{figure}
%TC:endignore

%% LIM
The Linear Inverse Model (LIM), first introduced by \citet{penland1995}, offers an alternative data-driven approach for ENSO forecasting. The LIM is a stochastic climate model \citep{hasselmann1976} that represents chaotic nonlinear dynamics by a deterministic multivariate linear system driven by noise that is white in time but may be correlated in space. This approximation may be suitable for dynamical systems with different components evolving on very different time scales, such as when the slowly-varying ocean is driven by and coupled to a much more rapidly decorrelating atmosphere \citep{frankignoul1977,penland1996}. Even when determined from anomaly covariances drawn from the short observational record alone, the LIM demonstrates seasonal-to-interannual tropical SSTA forecast skill on par with operational numerical prediction models \citep{newman2017,lheureux2020}. Refinements to the LIM, such as the cyclostationary (CS)-LIM to account for the annual cycle \citep{shin2021,vimont2022} and the inclusion of ocean memory \citep{newman2011,chen2016}, can further improve its predictive skill. 

% Problem Statement
The LIM generates climate variability that has a multivariate Gaussian distribution. The tropical Pacific SSTA distribution, however, is non-Gaussian and asymmetric, with warm events being more intense than cold events, while cold events are typically of longer duration \citep{takahashi2011,takahashi2016,okumura2019,geng2022}. As a correction to the LIM, we might first try modifying its stochastic forcing to include a linearly state-dependent noise term that is correlated with the original state-independent noise \citep{sardeshmukh2015}, the details of which in principle might also be extracted from observations \citep{martinez-villalobos2018}. Then, while the LIM’s predictable dynamics would remain linear, this additional “correlated additive-multiplicative” (CAM) noise term \citep{sardeshmukh2015a} would drive non-Gaussian variability including some key aspects of the observed asymmetry in the intensity and duration of ENSO events \citep{martinez-villalobos2019}. However, ENSO asymmetry has also been attributed to slower nonlinear processes in the tropical Pacific \citep{dommenget2013,takahashi2011,takahashi2019,hayashi2020}, which might not be simply represented by CAM noise. Given that the LIM already shows substantial predictive skill, then, we could systematically investigate its forecast error residual for predictable nonlinear dynamics, and add the resulting correction of the LIM. Developing such a hybrid LIM model is our aim in this study.

%% Hybrid models
Hybrid models, which combine numerical or empirical models with neural networks, offer a promising approach to capture unresolved dynamics in the low-data regime \citep{irrgang2021}. Instead of learning the complex system dynamics of the full system with a neural network, hybrid approaches are often more data efficient due to their simpler learning objective. Applications include post-processing of weather forecasts \citep{gneiting2005,bauer2015} and machine learning parametrization in coupled ocean-atmosphere models \citep{rasp2018,watt-meyer2021,kochkov2023}. 

In the context of ENSO forecasting, \citet{goel2017} and others \citep{wang2021,zhou2022} combined RNNs with statistical models for forecasting Niño indices. Expanding beyond ENSO index prediction, \citet{rodrigues2021} developed a hybrid model using the LIM operator within a ResNet-like architecture for global SSTA prediction. These hybrid approaches have not yet achieved state-of-the-art forecasting skill, and their potential advantages over fully deep learning models, such as interpretability and data efficiency, remain unexplored.

% Contribution
Here, we propose a hybrid deep-learning model for ENSO forecasting that combines the LIM with a Long-Short Term Memory (LSTM) network \citep{hochreiter1997}. This LIM-LSTM hybrid model is designed to capture the residuals between LIM forecasts and target data, thereby improving seasonal forecast accuracy. We diverge from existing methodologies by including seasonality in both the LIM and the LSTM. Furthermore, we adapt both the hybrid model and the LSTM to generate probabilistic ensemble forecasts of the full field variables, SSTA and SSHA. This is achieved by employing a set of output layers that generate ensemble members, designed to match the nonparametric distribution of the target data \citep{lessig2023}. We conduct a comparative analysis between our LIM-LSTM hybrid model and our LSTM deep-learning model, focusing on how each captures ENSO dynamics. 
\begin{figure}[!b]
    \centering
    \includegraphics[width=1.0\textwidth]{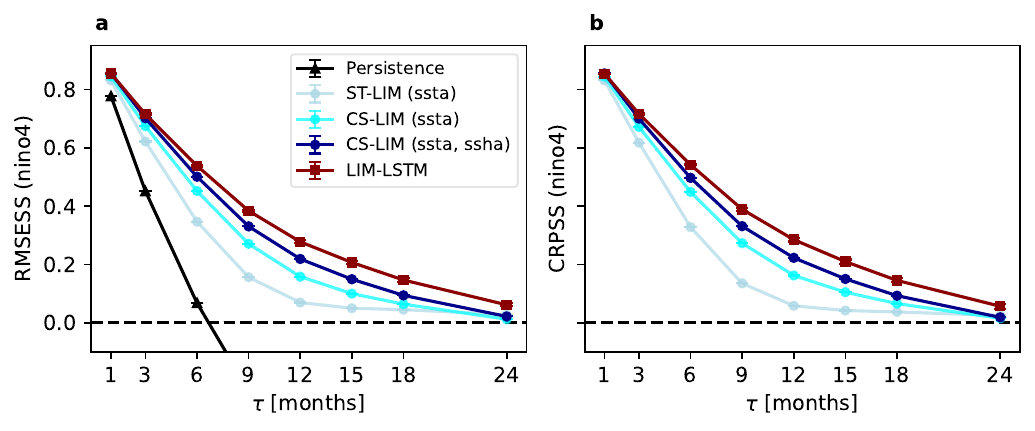}
    \caption{\textbf{RMSE and CRPS skill scores of the LIM versions and our LIM-LSTM model.} Skill scores for RMSE (\textbf{a}) and CRPS (\textbf{b}) across various LIM versions and the LIM-LSTM model are evaluated over forecast lead time ($\tau$) using the average SSTA in the Niño4 region on the test set. The progression in LIM versions from the stationary (ST)-LIM, which uses SSTA data and does not include seasonally-varying operators, to the more advanced cyclostationary (CS)-LIM incorporating seasonality, and then to the CS-LIM (ssta, ssha) that includes both seasonality and SSH factors is depicted. Enhancing the CS-LIM (ssta, ssha), our LIM-LSTM model utilizes an LSTM to effectively learn and adjust for its residuals.}
    \label{fig:skill_lims}
\end{figure}

One key question we address in this study is the training data requirements needed for each technique to achieve a given level of ENSO prediction skill. Therefore, we develop the technique on a large model dataset. While this has limitations, it allows us to compare the data requirements for both our hybrid and full DL models by evaluating their skill on subsets of the data with varying lengths. As an example, Fig.~\ref{fig:training_data} shows that for training on 50-100 years of monthly data, comparable in length to reanalysis products like ERA5 and CERA-20C, the 12-month forecast of both the CS-LIM and the LIM-LSTM hybrid model exhibit an Anomaly Correlation Coefficient (ACC) of bigger than 0.5 (the typical threshold for a forecast to be considered skillful). In contrast, for the same amount of training data, the 12-month forecast of the pure LSTM model’s has an ACC lower than 0.4, both evaluated on the 200-year long test set. It takes 300 years of data before the LSTM model reaches a forecast skill on par with the CS-LIM and LIM-LSTM hybrid model. Only when the training dataset size exceeds 500 years do both the LSTM and the LIM-LSTM model surpass the CS-LIM, with skillful forecasts (ACC > 0.5) up to leads of 18 months.
\section{Results} \label{sec:results}
\subsection{Improved forecast skill due to predictable nonlinearities}
  \label{sec:skill_improvement}
The CS-LIM estimated from the PCs of SSTA and SSHA of the first 1500 years of a CESM2 pre-industrial simulation in the tropical Pacific (see Methods sec.~\ref{sec:data}) captures the predictable linear dynamics of the system. We then trained an LSTM to learn the residuals between the CS-LIM’s predictions and the target data and combined these to form our hybrid LIM-LSTM model (see Methods sec.~\ref{sec:hybrid_model}). Improved forecast skill of the LIM-LSTM, relative to the LIM itself, can thus be attributed to its ability to capture predictable nonlinearities of the tropical Pacific Ocean dynamics. We quantified this improvement in skill by computing deterministic and probabilistic skill scores (RMSE and CRPS; see Methods sec.~\ref{sec:evaluation_metrics}) of the CS-LIM and LIM-LSTM on the test data (year 1800 to 2000). Both skill scores are obtained with respect to monthly climatology as a baseline model. 
\begin{figure}[!t]
    \centering
    \includegraphics[width=1.0\textwidth]{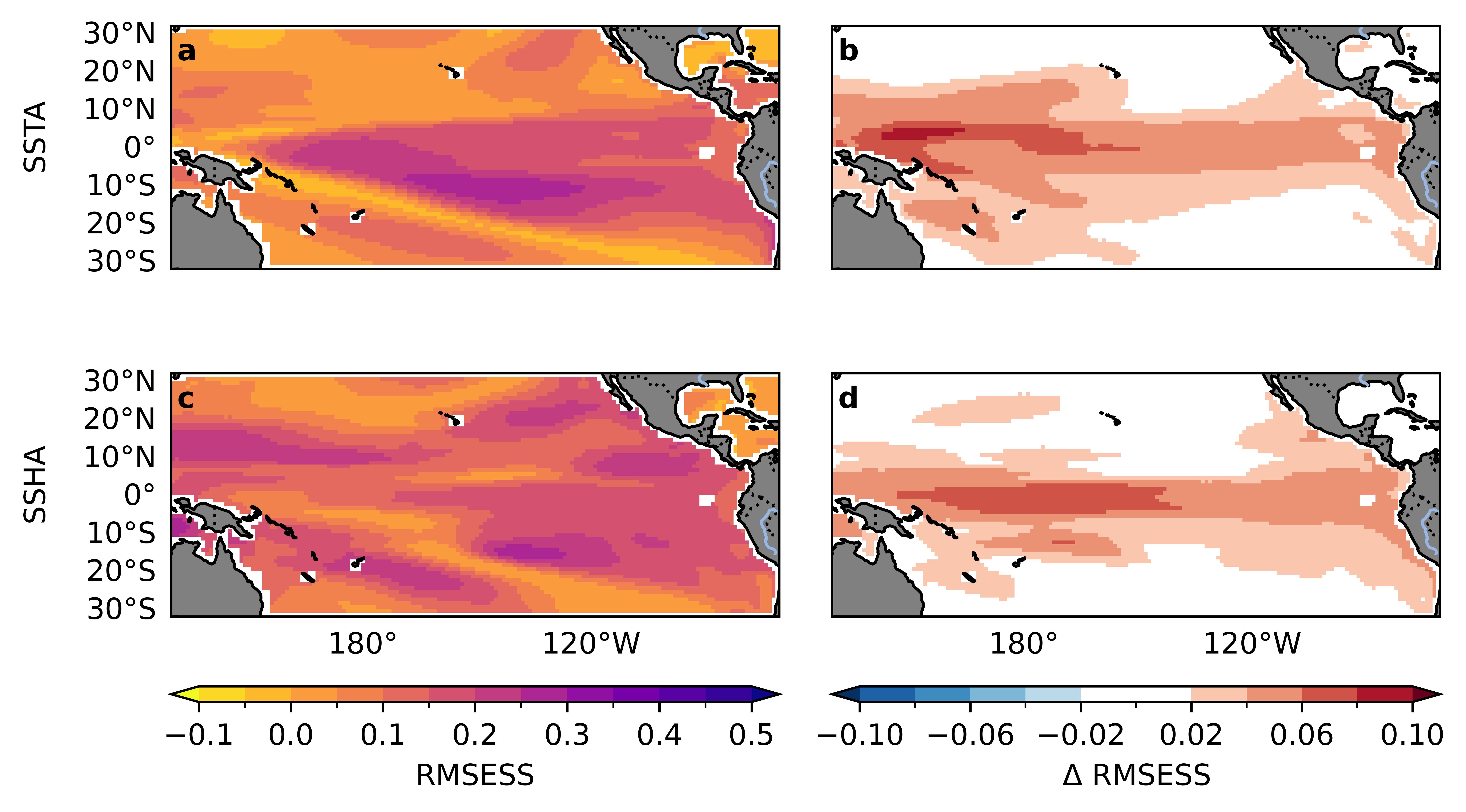}
    \caption{\textbf{Spatial distribution of skill improvement.} RMSE skill score of SSTA and SSHA for the $\tau=12$ month forecast of CS-LIM (\textbf{a}, \textbf{c}) and the differences in RMSE skill scores relative to the Hybrid model (\textbf{b}, \textbf{d}). Red colors indicate an improvement in skill in the LIM-LSTM model, while blue colors indicate a decrease in skill. Using a two-sided t-test, we evaluate the significance of the difference between the 1000 randomly bootstrapped means of CS-LIM, and the 95\% confidence interval threshold are shown.}
    \label{fig:skill_maps}
\end{figure}

First, we would like to capture as much of the linear dynamics in the LIM as possible, so that the LSTM is primarily tasked with learning nonlinear dynamics from the LIM errors. Various LIM variants, with increasing levels of complexity, are constructed to guide the choice of the best linear model. The initial LIM variant, formulated by \citet{penland1995}, is solely fitted to the first 30 PCs of SSTA in the Pacific and is termed stationary (ST)-LIM (ssta). An advancement to this is the CS-LIM (ssta), introduced by \citet{shin2021}, which includes seasonal variations and substantially surpasses the skill of the ST-LIM (ssta), as shown by the RMSE (Fig.~\ref{fig:skill_lims}a) and CRPS (Fig.~\ref{fig:skill_lims}b) skill scores of the Niño4 index. For reference, we present the skill of the persistence forecast which is worse than a climatological forecast (dashed line at zero) after $\tau$=6 months forecast lead time. Including a measure of ocean memory, by incorporating the first 10 PCs of SSHA in the tropical Pacific in the CS-LIM (ssta, ssha), leads to additional skill enhancement relative to the CS-LIM (ssta). We select SSHA for its availability in both model and observations, and for its strong relationship with important ENSO precursors, namely the upper ocean heat content and thermocline depth, following previous LIM studies \citep{johnson2001,newman2011}. Progressive improvements are evident in each LIM version, with the inclusion of seasonality and subsurface ocean variables contributing significantly to enhanced skill (Fig.~\ref{fig:skill_lims}). The CS-LIM (ssta, ssha) (CS-LIM hereafter) has the highest skill among the linear models and thus is considered the best representation of the predictable linear dynamics.
\begin{figure}[!t]
    \centering
    \includegraphics[width=1.0\textwidth]{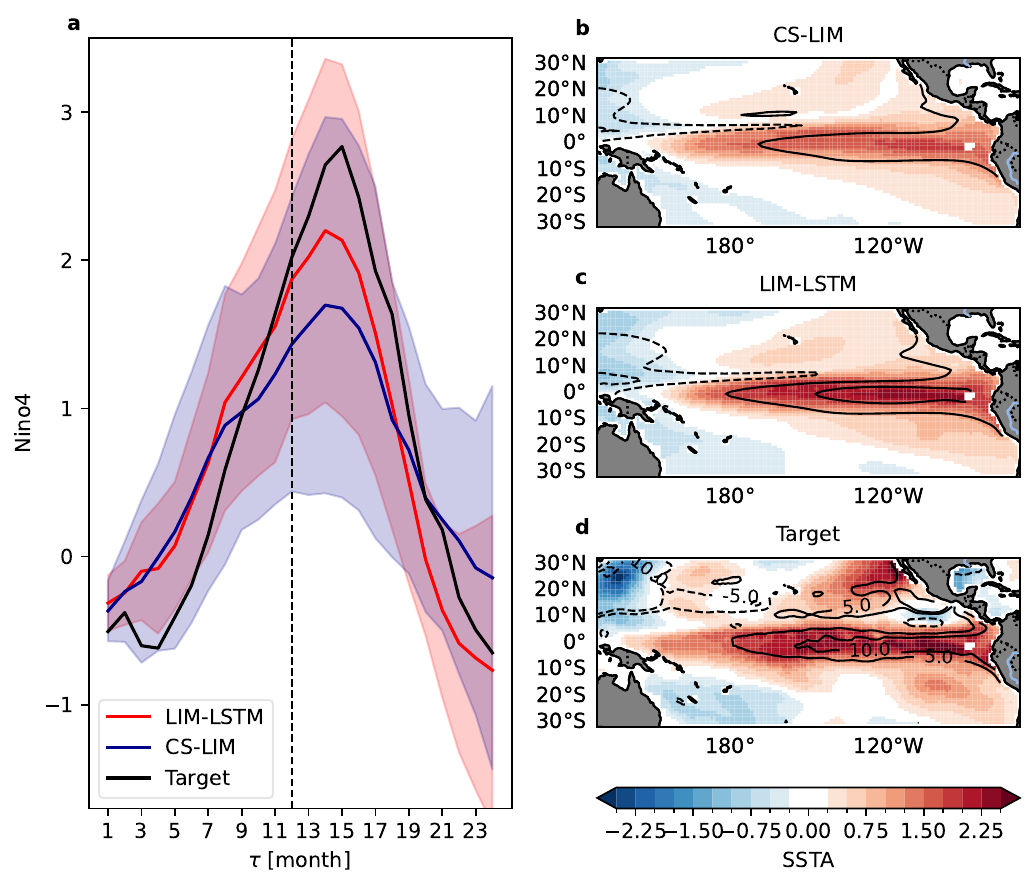}
    \caption{\textbf{Example El Niño forecast:} Example of a forecast initialized 12 months prior to an El Niño event exemplar. The Niño4 mean (solid line) and spread (shading) of the 16 ensemble members of the CS-LIM and LIM-LSTM forecast are shown in \textbf{a}. The dashed line in \textbf{a} indicates the 12-month lead. The mean SSTA forecast at $\tau=12$ months for the CS-LIM (\textbf{b}), LIM-LSTM (\textbf{b}), and target (\textbf{c}) are shown as color shadings while mean SSHA are depicted as contour lines.}
    \label{fig:example_en}
\end{figure}
Our hybrid LIM-LSTM model builds upon the CS-LIM by using an LSTM that learns to correct the error between the LIM forecast and the target data. The LSTM takes 16 ensemble member forecasts of the CS-LIM as input and learns to model their residual errors to the target data by minimizing the CRPS loss function detailed in Eq.~\ref{eq:crps-loss}. To ensure robustness, we repeat the model training five times with varied weight initialization and data shuffling, whose variability is depicted through error bars in Fig.~\ref{fig:skill_lims}. The skill improvement of the LIM-LSTM upon the CS-LIM is significant at forecast lead times larger than 6 months. We anticipate that the skill improvements can be largely attributed to predictable nonlinearities to the extent that we have tried to ensure that all known linear dynamics are captured by the CS-LIM.

We conducted a further examination of the seasonal dependency (Sec.~\ref{SI-sec:seasonal_skill}) and spatial distribution of skill for both the LIM and the LIM-LSTM model. The ensemble mean RMSESS for a 12-month CS-LIM forecast of SSTA exhibits the highest skill south of the equator and the Niño4 region while the skill is notably smaller in the upwelling region in the eastern tropical Pacific (Fig.~\ref{fig:skill_maps}a). In contrast, the RMSESS of SSHA demonstrated high skill in this region as well as in western tropical Pacific north of the equator (Fig.~\ref{fig:skill_maps}c) where the largest centers of SSH variability associated with ENSO occur \citep{capotondi2020}. 

The LIM-LSTM improves upon the CS-LIM skill throughout the Tropics (Fig.~\ref{fig:skill_maps}b, d). There is a discernible skill improvement all along the equator, with the most significant enhancement observed in the western tropical Pacific. This improvement is consistent with the patterns obtained using the CRPSS (not shown). The pattern of skill improvement is different from the pattern of CS-LIM skill; that is, the LIM-LSTM often has less improvement in regions where the CS-LIM already shows substantial skill. It is interesting to note that the western warm pool shows the highest skill improvement, and also exhibits the warmest temperatures in the Pacific where the nonlinear ocean-atmosphere feedback is the largest \citep{thual2023}. CESM2 and other climate models overestimate the SSTA variability in this region compared to observations which could increase this effect \citep{capotondi2020}. We defer a detailed analysis of nonlinear drivers learned by the LSTM to a future study.

\begin{figure}[!b]
    \centering
    \includegraphics[width=1.0\textwidth]{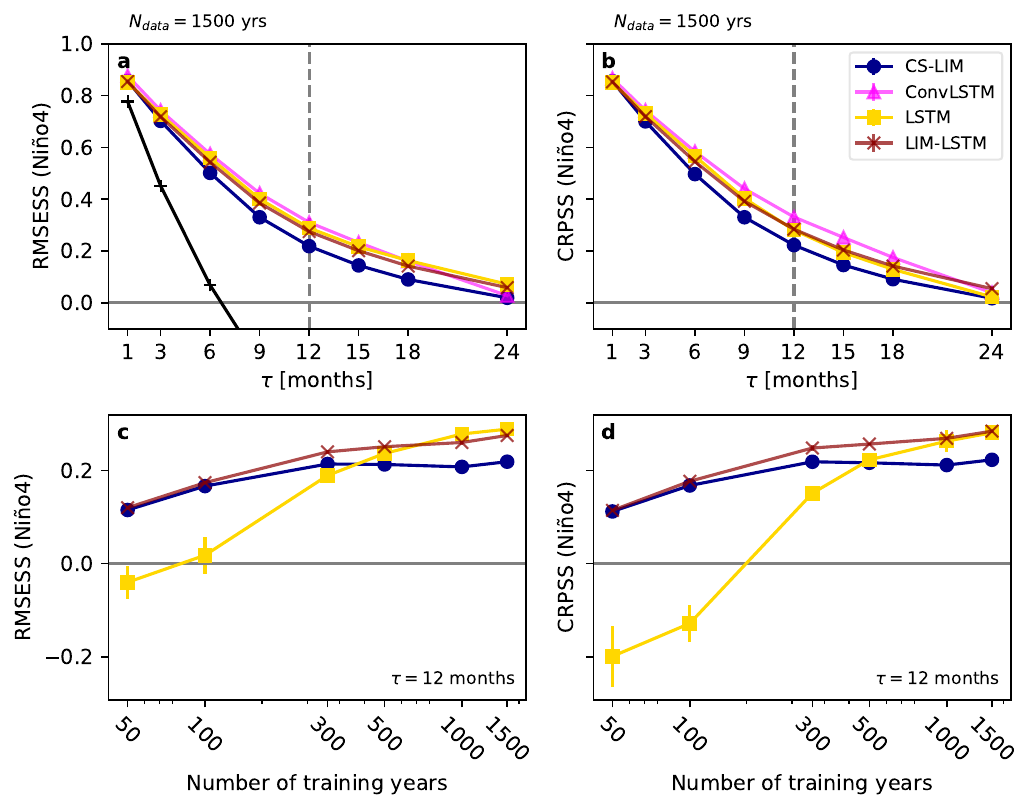}
    \caption{\textbf{Skill of deep learning baselines.} 
    Same as Fig.~\ref{fig:skill_lims} for the CS-LIM, LIM-LSTM, LSTM, and ConvLSTM trained on the 1500-year training set (\textbf{a}, \textbf{b}). The RMSESS and CRPSS ($\tau=12$ month forecast) of the models when trained on random subsets of the training set with 50 to 1500 years of data are shown in \textbf{c} and \textbf{d}. Skill scores are based on monthly climatology (Method Sec.~\ref{sec:evaluation_metrics}) with error bars reflecting model training runs with varied weight initialization and data shuffling.}
    \label{fig:skill_baselines}
\end{figure}

While both the LIM and LIM-LSTM can predict large El Niño events, the LIM-LSTM better captures event amplitudes. As an example, Fig.~\ref{fig:example_en} shows 24-month forecasts made by both models for a chosen El Niño exemplar drawn from the test dataset. The forecasts are initialized in December, 12 months before the peak of the chosen El Niño event (dashed line in Fig.~\ref{fig:example_en}A). Throughout this work, we show the Niño4 instead of the Niño3.4 region because of the CESM2 tendency to achieve the largest ENSO SST anomalies in the central Pacific rather than the eastern Pacific \citep{capotondi2020}. Results for the Niño3.4 index are qualitatively similar. Both the CS-LIM (red line) and LIM-LSTM (blue line) forecasts predict the observed warming, as depicted by the ensemble members’ mean, with the shading indicating their respective ensemble standard deviations. Notably, the LIM-LSTM forecast magnitude is closer to the CESM2 target data than the CS-LIM forecast. The LIM-LSTM’s uncertainty range, given by the standard deviation between 16 ensemble member forecasts, is also narrower than that of the CS-LIM forecast, yet still encompasses the target data, possibly suggesting a more accurate ensemble spread, consistent with the CRPSS results in Fig.~\ref{fig:skill_lims}. The SSTA and SSHA field forecasts at $\tau$=12 months lead time (Fig.~\ref{fig:example_en}C, D) show that the LIM-LSTM generally better captures this El Niño event's magnitude throughout the equatorial Pacific. However, both the CS-LIM and LIM-LSTM lack some smaller spatial structures and extratropical features evident in the verification fields. 
\subsection{Comparison to deep learning baselines}
We conducted a comparative analysis between our hybrid LIM-LSTM and two fully deep learning models, the LSTM and the ConvLSTM (see Methods Sec.~\ref{sec:materials-methods}). While the LSTM produces forecasts in the PC space of SSTA and SSHA, the ConvLSTM takes the field variables in the tropical Pacific as input and forecasts them directly. We employ both models to ensure that the EOF truncation does not substantially hamper our forecast skill. Similar to the training of the LIM-LSTM, each model underwent five separate training sessions with varied weight initialization and data shuffling (error bars in Fig.~\ref{fig:skill_baselines}). The deep learning models have RMSE and CPRS skill scores that are similar to the LIM-LSTM model (Fig.~\ref{fig:skill_baselines}a, b), though the ConvLSTM exhibits a slight improvement at the 12-month forecast horizon. This suggests that the LIM-LSTM model successfully captures most of the predictable dynamics, with the marginal gains of the ConvLSTM likely attributable to the PC truncation of our hybrid model. Crucially, the LIM-LSTM model achieves this level of forecasting skill with significantly fewer parameters compared to the full deep learning models.
% TODO: number of model parameters 
This aspect is particularly beneficial in scenarios with limited data, as shown in Fig.~\ref{fig:skill_baselines}c and d. With less than 500 years of monthly data, both the LIM and LIM-LSTM exhibit higher 12-month RMSES and CRPS skill scores than the LSTM. 

\subsection{Predictability assessment in the LIM-LSTM model}

\begin{figure}[!t]
    \centering
    \includegraphics[width=1.0\textwidth]{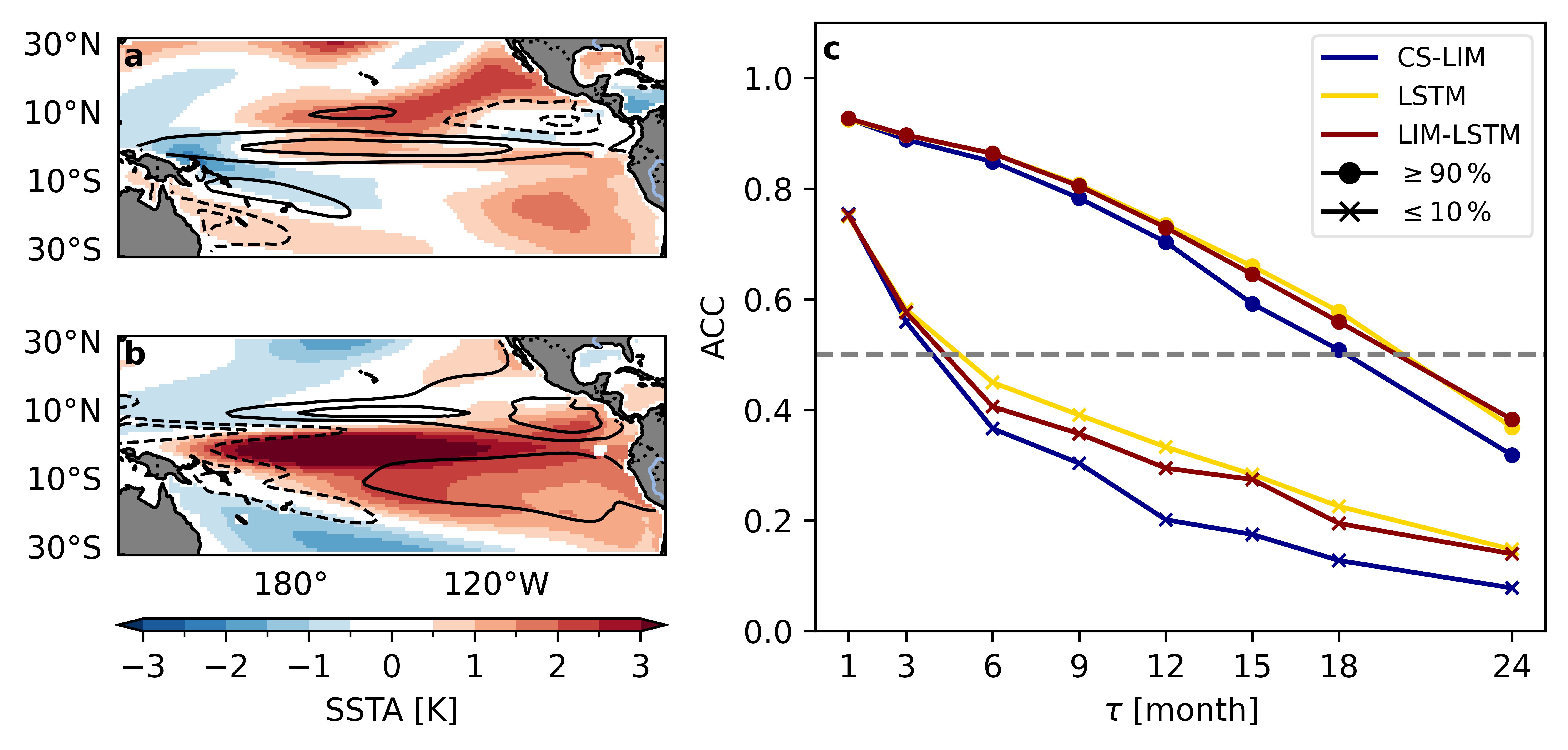}
    \caption{\textbf{Predictability is captured by linear optimals.} The optimal initial condition (OIC) (a) of the CS-LIM for a 12-month lead forecast initialized in April evolves into an El Niño-like pattern after 12 months (b). Forecasts are initialized from states in the test set whose projection on the OIC have the least (0-10\%) or greatest (90-100\%) amplitude. The anomaly correlation coefficient (ACC) of these forecasts are shown for the CS-LIM, LIM-LSTM and LSTM (c). ACC curves initialized from the OICs with the smallest amplitude correspond to x's whereas those initialized from OICs with the largest amplitude correspond to o's.}
    \label{fig:theoretical_skill}
\end{figure}

Previous LIM predictability studies have shown that the LIM can predict not only its own skill but, often, also that of operational prediction models \citep{newman2017,albers2019,albers2021}. We might exploit this ability by building our hybrid LIM-LSTM model on top of the LIM. For any given initial forecast state, the CS-LIM's maximum predictability can be calculated by projecting the state onto the optimal initial condition (OIC) for a forecast lead time $\tau$. The OIC is the singular vector corresponding to the largest singular value of the forecast propagator $G$(see Methods \ref{sec:LIM}). This condition is optimal in the sense that of all possible initial conditions of unit amplitude, it evolves into the largest amplitude (under the L2 norm) state vector at lead time $\tau$ \citep{penland1995}. For CS-LIM, we obtain a different OIC at every month and lead time.

The CS-LIM's OIC for a 12-month forecast in April (Fig.~\ref{fig:theoretical_skill}a) exhibits an SST and SSH structure that aligns closely with earlier research \citep{newman2011,vimont2014,capotondi2015a}. Key features of the OIC are: a band of positive SST anomalies in the northern subtropics, stretching diagonally from approximately 0°N, 180°W northeastward to around 30°N, 120°W; positive SST anomalies in the southern subtropics, predominantly east of 120°W; enhanced thermocline depth anomalies along the equator; and comparatively weak negative thermocline depth anomalies located at approximately 10°N in the eastern tropical Pacific. We obtain its subsequent evolution after $\tau=12$ months by applying the LIM operator to the OIC (Fig~\ref{fig:theoretical_skill}b). It is important to note that the magnitude of these patterns is arbitrary, a result of the unit norm of the singular vector.

The strength of the projection of the initial forecast state onto the CS-LIM $\tau$-lead OIC, or how well the initial state matches the first singular vector – termed as optimal initial growth – is a key determinant of the potential forecast skill of the CS-LIM \citep{newman2003}. To illustrate this point, for each forecast lead time ranging from 1-24 months, we select initial states from the test set whose projections on the OIC determined for that lead have either the least (0-10\%) or greatest (90-100\%) amplitudes, and then compute the forecast skill that results from forecasts made for each group separately. The average ACC, illustrated in Fig.~\ref{fig:theoretical_skill}c, is substantially larger for forecasts initialized from states with the strongest optimal initial growth than for states with minimal optimal initial growth, a result we could have anticipated at the time of forecast.

The dependence of skill upon the presence of (linear) optimal initial growth occurs not only for the CS-LIM but also for the LIM-LSTM  hybrid model, although it also has a substantial skill increase compared to the CS-LIM (Fig.~\ref{fig:theoretical_skill}c). This result implies that optimal initial growth, although a property derived from the CS-LIM, influences not only linear predictability but also significantly impacts overall predictability of the entire system. This is further supported by the LSTM forecast, where a similar marked increase in skill is observed for initial states with the strongest optimal initial growth (i.e., cases where strong linear predictability is expected) as opposed to those with minimal initial growth.

\subsection{ENSO asymmetry is nonlinearly predictable}

\begin{figure}[!t]
    \centering
    \includegraphics[width=1.0\textwidth]{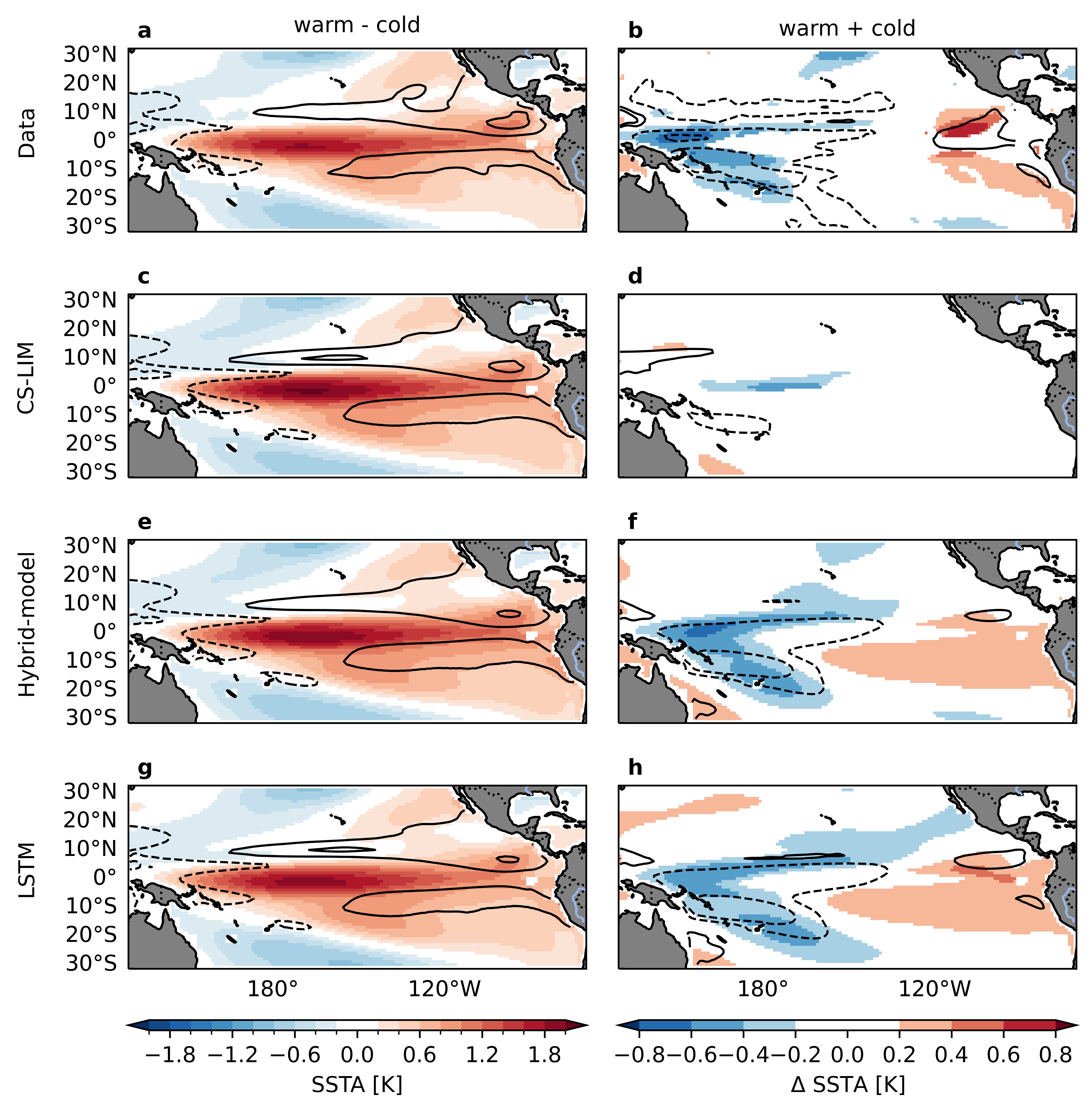}
    \caption{\textbf{Nonlinear models capture ENSO asymmetries.} The 12-month evolutions of states initialized in April with the absolute largest optimal initial growth (>90\%) show warm and cold patterns. The average target state of warm-cold patterns (a) and warm+cold patterns (b) are depicted, as well as their forecasts of the CS-LIM (c, d), LIM-LSTM model (e, f) and LSTM (g, h). 
    We use a two-sided t-test to evaluate the significance of the difference between the means of the cold and the warm patterns. Only those values that surpass the 95\% confidence interval threshold are shown.}
    \label{fig:projection_optimals}
\end{figure}

The initial states projected onto the OIC, shown in \ref{fig:theoretical_skill}a, can yield either positive or negative optimal initial growth which evolve into warm or cold patterns, respectively. We examine the average $\tau=12$ month forecast of April initial states with the absolute largest optimal initial growth (>90\%) for our CS-LIM, LIM-LSTM, and LSTM models. 

The average forecast pattern’s magnitude and asymmetry are estimated by warm minus cold patterns (W-C, Fig.~\ref{fig:projection_optimals}, first column) and warm plus cold patterns (W+C, Fig.~\ref{fig:projection_optimals}, second column), respectively. A two-sided t-test is utilized to ascertain statistically significant difference between the means of the two distributions. We report only those values that surpass the 95\% confidence interval threshold. 

The W-C pattern in the target data (Fig.~\ref{fig:projection_optimals}a) closely resembles the evolved optimal pattern of the CS-LIM (Fig.~\ref{fig:theoretical_skill}b) and its average empirical forecast (Fig.~\ref{fig:projection_optimals}c). An evident east-west dipole structure is observed in the asymmetry of warm versus cold events of the target data (Fig.~\ref{fig:projection_optimals}b). Cold events exhibit higher SSTA and SSHA magnitudes in the western tropical Pacific, while warm events show greater magnitudes in the Central-Eastern Pacific. This asymmetry is not present in the W+C patterns of the CS-LIM forecast (Fig.~\ref{fig:projection_optimals}d) because of its linear and therefore symmetric evolution of cold and warm events. The remaining subtle differences between warm and cold patterns of CS-LIM forecast likely originate from the asymmetrical distribution of initial conditions, which might arise by chance but might also arise from a CAM noise process (not included in our CS-LIM), as noted earlier.

In contrast, the LIM-LSTM model and LSTM forecasts accurately capture both the magnitude and asymmetry of warm and cold events. Both nonlinear model forecasts exhibit the zonal dipole structure present in the target data (Fig.~\ref{fig:projection_optimals}f and h). This finding highlights the ability of our nonlinear models to predict the asymmetry between warm and cold patterns. Note also that the predicted W-C component for both the LIM-LSTM and LSTM models is very similar to that of the CS-LIM (see Fig.~\ref{fig:projection_optimals}c,e,g). This suggests that the hybrid model improves upon the CS-LIM by capturing predictable nonlinearity, rather than by finding additional linear skill. Taken together, these findings underscore the LIM-LSTM model's potential to disentangle linear and nonlinear predictable dynamics, setting the stage for future systematic analysis of nonlinearities in subsequent work.

\section{Discussion} \label{sec:discussion}

We have introduced a LIM-LSTM hybrid model specifically tailored for forecasting SST and SSH anomalies in the tropical Pacific. We start from the LIM, an empirical model describing the dynamics of the slower-varying ocean as stochastically forced by the rapidly varying atmosphere, with its deterministic dynamics assumed to be linear. However, while the LIM produces ENSO forecasts comparable to state-of-the-art numerical models, it is unable to capture observed asymmetries of ENSO that are likely important to its predictability, especially for longer forecast leads.

We combine an LSTM with the LIM to capture predictable nonlinearities and non-Markovian dynamics in the evolution of monthly tropical SSTA. The LIM-LSTM model is trained and tested on SSTA and SSHA data from the CESM2 pre-industrial control run. We find that modeling nonlinearities significantly enhances the forecast accuracy, particularly in the western tropical Pacific within the 9 to 18-month range.

% Sources for nonlinearity
Our findings provide initial evidence that the asymmetry between warm and cold events is a key source of nonlinearity that improves forecasting skill beyond linear models. This first insight lays the groundwork for a more comprehensive follow-up characterization of the potential sources of nonlinearity of ENSO forecasting.

% Asses of predictability
We demonstrate that the predictability of the LIM-LSTM model is strongly related to the theoretical expected skill of the LIM which allows us to reliably assess its predictability. In contrast, neural networks typically struggle to provide accurate predictability assessments on seasonal to annual scales, primarily hindered by their weak spread-to-skill relationship. Given that LIM models show competitive performance with GCMs \citep{newman2017,albers2019, albers2021} and that the LIM-LSTM requires less data than a full LSTM, this hybrid LIM-LSTM is a promising approach for other S2S climate forecasts beyond ENSO.

% Lower data requirement for training
A notable feature of our LIM-LSTM model is its data efficiency, particularly when compared to more data-intensive deep learning models like the LSTM network. This aspect is crucial given the limited span of available oceanic observational data. For a fair comparison, we utilized data from GCMs in our training, acknowledging their inherent biases as discussed in our study. The use of domain adaptation methods from deep learning emerges as a promising strategy to close the gap between GCM data and observational data. However, the field still needs more research to fully understand how neural network models can be adjusted to observational data when pre-trained on simulated data.

Although our models are based on LSTMs, which are not the most advanced deep learning models, they exhibit similar forecasting skill to the larger vision transformer model by Zhou and Zhang \citep{zhou2023}. This suggests that for the amount of observational data typically available to train data-driven models on seasonal time scales, hybrid models could outperform fully deep-learning models.

%TC:ignore
\newpage
\section{Materials and Methods} \label{sec:materials-methods}

\subsection{Data} \label{sec:data}

Training DL models for ENSO prediction with monthly data is limited by the short observational record. To test the impact of record length on the data-driven models, we use the 2000-year CESM2 pre-industrial control simulation \citep{danabasoglu2020}. We focus on monthly SST and SSH data in the tropical Pacific region (130°E - 70°W, 31°S - 32°N), which we linearly interpolate to a resolution of 1°x1°. SSH is a proxy for the upper ocean temperatures and the thermocline depth. Despite the lack of external forcing in the control simulation, we observe a trend in SST data. We linearly detrend the data and remove the seasonal cycle by subtracting the monthly climatology which is obtained over the training set (0-1500). As SSTA and SSHA differ in units and scales, we perform a $z$-score normalization before model training. Both the LIM and LIM-LSTM model require us to reduce the dimensionality, which is achieved using Empirical Orthogonal Function (EOF) analysis. The dataset is divided into training (75\%, year 1-1500), validation (15\%, years 1500-1800), and test set (10\%, 1800-2000), where the validation set is used for refining the hyperparameters of our models.

\subsection{Methods} \label{sec:methods}

The objective of our study is to accurately predict SSTA and SSHA fields for a specified forecast lead time $\tau$. We define the stacked variable fields at a given time $t$ as $\mathbf{x}(t) = (\mathbf{x}_\text{SSTA}, \mathbf{x}_\text{SSHA}) (t)$, where each field spans the tropical Pacific $\mathbf{x}_\text{SSTA/SSHA} (t) \in \mathbb{R}^{N_{lat} \times N_{lon}}$. We estimate a function $f_{m(t)}$, depending on month $m(t)$, that predicts the future state, $\hat{\mathbf{x}}(t+\delta)$ at an incremental time step $\delta$, by 

\begin{equation} 
    \mathbf{\hat{x}}_i (t+\delta) = f_{m(t)} \left( \mathbf{x}(t), f_{m(t-\delta)} \left( \mathbf{x}(t-\delta), \ldots, f_{m(t-t_\text{hist})} \left( \mathbf{x}(t-t_\text{hist}) \right) \right) \right),
    \label{eq:autoregressive}
\end{equation} 

The autoregressive prediction is based on the previous states, $\mathbf{x}(t-\delta), \ldots, \mathbf{x}(t-t_\text{hist})$, referred to as history. All our model forecasts consist of $i$ ensemble members, which allows us to estimate the model uncertainty. The dynamics of the tropical Pacific Ocean show a strong seasonal phase locking. We therefore condition $f$ on the month of the year, $m(t)$. The month conditioning depends on the model architecture and will be discussed for each model separately.

\subsection{Empirical Orthogonal Function (EOF)} \label{sec:EOF}

Estimating the linear evolution operator of the LIM requires a matrix inversion of the number of input dimensions. The matrix inversion is intractable for the full spatial fields of SSTA and SSHA. For this reason, each state $\mathbf{x}(t)$ is transformed into a lower-dimensional state $\mathbf{z}(t) = \left( \mathbf{z}_\text{SSTA}, \mathbf{z}_\text{SSHA} \right)$. Dimensionality reduction of the SSTA and SSHA fields in the tropical Pacific is achieved through EOF analysis, utilizing the first 20 Principal Components (PCs) for SSTA and the first 10 PCs for SSHA. Including higher-order PCs does not affect our results. Forecasting in the lower dimensional space is then equivalently conducted on these PCs, as formulated in Eq.~\ref{eq:autoregressive}. For analysis and evaluation, we transform our forecast back to grid space. To adequately replicate the high spatial frequencies of the input fields, we add variability by randomly sampling loadings of the higher-order PCs (20-300) for both SSTA and SSHA at each time step. These random features are then combined with their respective EOFs and added to the forecast fields, ensuring a closer match to the spatial intricacies of the original data.

\subsection{Linear Inverse Model}
    \label{sec:LIM}

The LIM describes the dynamic of the tropical Pacific as a multivariate linear system subject to stochastic forcing from the atmosphere. The underlying dynamics of such a system is described by a linear stochastic differential equation,

\begin{equation}
    \frac{d\mathbf{z}}{dt} = \mathbf{L} \mathbf{z} + \mathbf{\zeta}
    \label{eq:SDE}
\end{equation}

where $\mathbf{L}$ is the linear operator describing the dynamics of $\mathbf{z}$ and $\mathbf{\zeta} \sim \mathcal{N}(0, \mathbf{Q})$ is a noise vector that is uncorrelated over time but spatially correlated, as encoded in the covariance matrix $\mathbf{Q}$. Forecasts of $\mathbf{z}$ for a lead time $\tau$ are given by the transition probability as, 

\begin{equation}
    p \left( \mathbf{z}(t+\tau) | \mathbf{z}(t) \right) = \mathcal{N} \left( \mathbf{z}(t+\tau); \mathbf{\mu}_\tau (t), \Sigma_\tau \right) 
    \label{eq:transition_probability}
\end{equation}
\begin{equation}
    \text{with} \qquad \mathbf{\mu}_\tau (t) := e^{\mathbf{L} \tau} \mathbf{z} (t) \qquad \text{and} \qquad \Sigma_\tau := \int_{0}^{\tau} e^{\mathbf{L}s} \mathbf{Q} \mathbf{Q}^T e^{\mathbf{L}^T s} ds,
    \label{eq:transition_probability_mean_cov}
\end{equation}

where $\mathbf{\mu}_\tau (t)$ is the infinite ensemble mean forecast and $\Sigma_\tau$ is the forecast covariance matrix.

\citet{penland1995} show that the linear operator $\mathbf{L}$ and the noise covariance $\mathbf{Q}$ can be estimated from the data under two assumptions. First, the system has to be statistically stationary which allows us to write the Fluctuation-Dissipation relationship as

\begin{equation}
    \mathbf{L} \mathbf{C}(0) + \mathbf{C}(0) \mathbf{L}^T + \mathbf{Q} = 0,
    \label{eq:fluctuation-dissipation-relationship}
\end{equation}

where $\mathbf{C}(0) = \langle \mathbf{z}(t) \mathbf{z}^T (t)\rangle$ is the spatial covariance matrix. Secondly, the autocorrelation of the system decays with lead time $\tau$ which can be expressed using the time-lead covariance matrix $\mathbf{C}(\tau) = \langle \mathbf{z}(t + \tau) \mathbf{z}^T (t)\rangle$ as

\begin{equation}
    \lim_{\tau \rightarrow \infty} \mathbf{C}(\tau) \mathbf{C}(0) = 0 \Rightarrow \mathbf{C}(\tau) = e^{\mathbf{L} \tau} \mathbf{C}(0),
    \label{eq:decay_autocorrelation}
\end{equation}

where $\mathbf{G}(\tau):= \exp(\mathbf{L} \tau)$ is the Greens function that must tend to zero for long lead times. Typically, both assumptions hold for detrended anomaly data of a chaotic system. 

Once $\mathbf{L}$ and $\mathbf{Q}$ are estimated from the data, we obtain forecast trajectories from an initial time $t$ to $t+T$, by numerically integrating eq. \ref{eq:SDE} using the forward Euler-method with incremental update steps $\delta$ as 

\begin{equation}
    \mathbf{z}(t+\delta) = \mathbf{z}(t) + \mathbf{L} \mathbf{z} (t) \delta + \zeta \sqrt{\delta}, 
    \label{eq:numerical_integration_lim}
\end{equation}

where $\zeta \sim \mathcal{N}\left( 0, \mathbf{Q} \right)$ is a random sample from the noise distribution. We create $n$ ensemble member trajectories from $t$ to $t+T$ when integrating the system $n$-times. The infinite ensemble member mean is given by $\mathbf{\mu}(t)$ in eq. \ref{eq:transition_probability}.

In the equatorial Pacific, the variance in SSTA shows a distinct annual pattern with low variance during the boreal spring and high variance in the boreal winter. This peak in winter variance aligns with the occurrence of the most intense warm and cold ENSO events, a phenomenon referred to as "ENSO phase locking" \citep{rasmusson1982}. \citet{shin2021} showed that including seasonality in the LIM improves its forecast reliability. Their cyclostationary (CS)-LIM involves estimating unique linear operators and noise covariances for each month, indexed using $m(t)=1,2, \ldots, 12$. The numerical integration of the stationary (ST)-LIM outlined in eq. \ref{eq:numerical_integration_lim} changes to 

\begin{equation}
    \mathbf{z}(t+\delta) = \mathbf{z}(t) + \mathbf{L}_{m(t)}^{CS} \mathbf{z}(t) \delta + \zeta^{CS}_{m(t)} \sqrt{\delta}, 
\end{equation}

where $\zeta^{CS}_j \sim \mathcal{N}\left( 0, \mathbf{Q}^{CS}_j \right)$. The CS-LIM forms the base model of our LIM-LSTM model outlined in the following.

\subsection{LIM-LSTM model}
  \label{sec:hybrid_model}

\begin{figure}[tp]
    \centering
    \includegraphics[width=1.\textwidth]{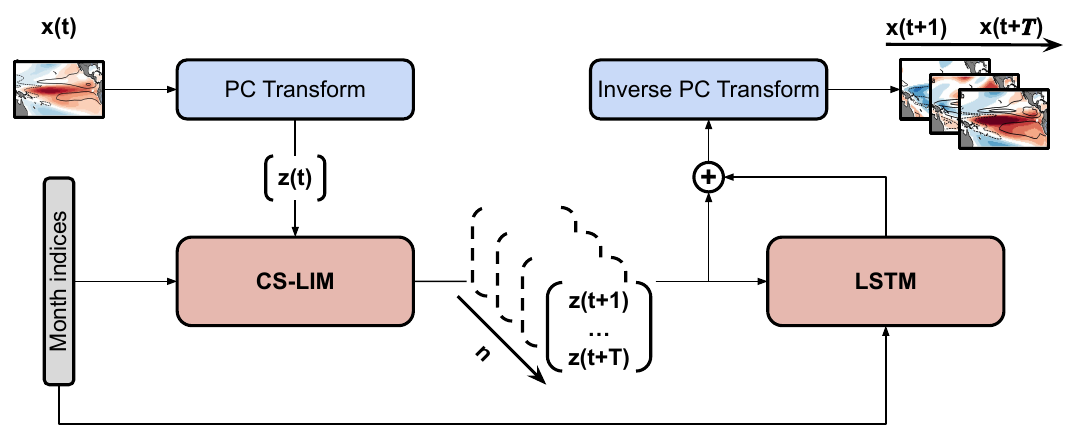}
    \caption{\textbf{Schematic Representation of the LIM-LSTM model}. 
    First, the initial state at time $t$ is projected onto the PCs. This is
    followed by an ensemble forecast using the CS-LIM which is conditioned on
    the forecast months. Subsequently, the LSTM adjusts each ensemble member of the linear
    CS-LIM forecast. Finally, the refined forecast is transformed
    back to grid space by multiplying the PCs with the respective EOF patterns.
    }
    \label{fig:architecturc_hybrid}
\end{figure}

We introduce a novel LIM-LSTM model that combines the LIM with an LSTM network. While the LIM captures the predictable linear dynamics, the LSTM learns the residuals between the LIM predictions and the actual data, thus the nonlinear dynamics. Our methodology is schematically detailed in Fig. \ref{fig:architecturc_hybrid}.

During inference, we project the initial state of the tropical Pacific, $\mathbf{x}(t)$, onto the leading EOFs and employ the LIM to predict future states over $\tau = 1, \ldots, T$ timesteps. For each timestep, $t+\tau$, we predict a correction, $\hat{\mathbf{z}}_\text{res}(t+\tau)$, to the LIM forecast, $\hat{\mathbf{z}}_\text{LIM}(t+\tau)$. The final forecast is thus defined as:

\begin{equation}
    \hat{\mathbf{z}}(t+\tau) = \hat{\mathbf{z}}_\text{LIM}(t+\tau) + \hat{\mathbf{z}}_\text{res}(t+\tau).
\end{equation}

The nonlinear correction is modeled by the LSTM as $\hat{\mathbf{z}}_\text{res}(t+\tau) = \text{LSTM} \left(\hat{\mathbf{z}}_\text{LIM}(t+\tau), \mathbf{h}(t) \right)$, where $\mathbf{h}(t)$ is the latent state of the LSTM which aggregates the information of previous states. The LSTM is selected not only for its ability to capture nonlinear relationships inherent in deep neural networks but also for learning non-Markovian dynamics.

To include seasonality within the LSTM, we introduce a learned affine transformation to its latent state \citep{perez2017}, $\mathbf{h}(t)$, as follows:

\begin{equation}
    \mathbf{h}_{m(t)}(t) = (1 + \alpha_{m(t)}) \mathbf{h}(t) + \beta_{m(t)},
    \label{eq:film}
\end{equation} 

where $\alpha_{m(t)}$ and $\beta_{m(t)}$ represent embeddings for each month, enabling the network to adapt its latent state dynamically to seasonal variations.

The LSTM component is configured to process the forecast from each CS-LIM ensemble member, represented as $\left[\hat{\mathbf{z}}(t+1),...,\hat{\mathbf{z}}(t+T) \right]$. At each timestep, the input to the LSTM network is linearly projected into a higher-dimensional latent space where it is processed by two consecutive LSTM layers. By combining the LIM forecast at each timestep with the LSTM's hidden state from the previous time step, the model iteratively accumulates information across the entire forecast sequence. Finally, the predicted latent states are linearly projected back onto the PCs and added to the original LIM prediction. The combined prediction is then transformed to grid space by multiplication with the respective EOFs, as described in sec.~\ref{sec:methods}\ref{sec:EOF}. 

\subsection{Purely deep learning baselines} \label{sec:dl_baselines}

To verify the utility of our LIM-LSTM model, we provide a comparison against fully neural network-based approaches. Similar to the structure of our LIM-LSTM model, we construct an LSTM that operates in the PC space. Additionally, we explore the application of a Convolutional LSTM (ConvLSTM \citep{shi2015}) architecture in grid space. We employ a custom ConvLSTM variant specifically tailored to perform forecasts on fields with large-scale spatial structures. 

Both the PC-LSTM and the ConvLSTM have a standard Encoder-Decoder structure used for sequence-to-sequence modeling \citep{sutskever2014}, as depicted in Fig.~\ref{SI-fig:architecture_lstm}. Unlike the LIM and LIM-LSTM model, these models incorporate information from time points preceding the initialization time. The Encoder network is designed to aggregate this historical information into a latent state which initializes the Decoder model. It begins with a downsampling block that transforms the history into a higher-dimensional latent space, which is then processed by two LSTM layers, where the input is added to the hidden state from the previous time step. The hidden state of the second LSTM layer at time $t$ is then passed to the Decoder network. The Decoder mirrors the Encoder with two LSTM layers of its own, which do not require any additional input other than the hidden state and can thus be rolled out over the full prediction horizon $t+T$, transferring the hidden state autoregressively for each successive prediction time. This is followed by an upsampling block that transforms the hidden state back to the input grid space. To generate 16-ensemble members, we employ a separate upsampling block for each member which are trained jointly. Similar to the LIM-LSTM model, we incorporate seasonal information through a learned affine transformation in both the Encoder and Decoder networks.

\paragraph*{PC-LSTM}

In the PC-LSTM, the downsampling consists of an EOF-truncation of the entire SSTA and SSHA fields onto their respective PCs (sec.~\ref{sec:methods}\ref{sec:EOF}), followed by a learned linear layer that projects the data into the latent space. The latent states are recursively predicted using LSTM layers, in both the encoder and the decoder network. Equivalently to the downsampling, the upsampling block starts with a linear layer to project the latent space back to the PCs, followed by a projectionon their respective EOFs to yield a forecast of the variables in grid space. 

\paragraph*{ConvLSTM}

We employ a second DL model that does not involve EOF-truncation of data. This model is based-on the ConvLSTM \citep{shi2015} with a customized LSTM cell (Fig.~\ref{SI-fig:architecture_swinlstm}). Unlike the PC-LSTM, the input, latent, and hidden states in our model maintain dimensions of channel, height, and width, albeit of varying sizes. For model input, both variables are stacked along the channel dimension. The encoder initially downscales the height and width dimensions to reduce computational costs, while simultaneously expanding the channel dimension via a strided convolutional layer. Following the methodology of \citet{liu2021, liu2022} we use equal stride and kernel sizes for this initial encoding, effectively partitioning the input into small patches of equal size (4x4 grid-steps in latitude/longitude direction). This method of dimensionality reduction differs from the EOF as it is learned end-to-end with the forecasting model, as well as being based on local patches, rather than global features as encoded in EOFs. Our approach further adapts the standard ConvLSTM layer, in a similar fashion to \citet{liu2022}, by splitting the convolution into spatial and channel mixing components, interspersed with layer normalizations, as depicted in Fig.\ref{SI-fig:architecture_swinlstm}. This modification facilitates two major improvements over standard ConvLSTMSs, (i) the spatial mixing component enables a substantially larger receptive field while reducing the number of parameters and (ii) the added normalization stabilizes training and supports conditioning on monthly embeddings, implemented through the affine transformation outlined in Eq.\ref{eq:film}. Finally, the mirrored Decoder network, which also consists of two ConvLSTM layers and a strided and transposed convolution, transforms the aggregated hidden state back into the full-resolution grid space. An ensemble of 16 separate final projection layers are used to generate a probabilistic prediction and the model is trained using the Ensemble CRPS as detailed below.

\subsection{Optimization procedure}

Each model is designed to produce a probabilistic prediction by generating an ensemble of 16 sequences. We train the models to enhance both the accuracy of individual predictions and the spread among ensemble members. This is achieved by employing the Continuous Ranked Probability Score (CRPS) for optimization. The CRPS is a probabilistic metric that compares the cumulative probability distribution (CDF) of the forecast to the CDF of the target. The target, $\mathbf{x}(t)$, is a single observation, and thus its CDF is a step function. A perfect CRPS score of zero would be a Dirac delta-like predictive probability density function centered at the target, for which the CDF would be the same step-function as is for the target. The CRPS has an analytic expression for parametric distributions, like the Gaussian distribution, but also a statistical form for empirical distributions \citep{hersbach2000, gneiting2007}.

For our $M$-ensemble member prediction, we compute the pixel-wise CRPS for empirical distributions as,

\begin{equation}
    \text{CRPS}(\hat{X_k}(t), x_k(t)) = E[|\hat{X_k}(t)- x_k(t)|]-\frac{1}{2} \cdot E\left[\left|\hat{X_k}(t)-\hat{X_k}^{\prime}(t)\right|\right],
    \label{eq:crps}
\end{equation}

where $\hat{X_k}(t)$ and $\hat{X_k}^{\prime}(t)$ are the $M$-ensemble member prediction and $x_k(t)$ the target value at each location $k = 1, \hdots, 2 \times N_{lat} \times N_{lon}$. We optimize the parameters of each model by minimizing the averaged CRPS between the observed data, $\mathbf{x}(t+\tau)$, and its ensemble forecast, $\hat{\mathbf{X}}(t+\tau)$, across all locations and lead $\tau \in [1, T]$. Using Eq.~\ref{eq:crps}, we define our tailored loss function as:

\begin{align}
    l \left( \hat{X}(t), \mathbf{x}(t) \right) &= \sum_{\tau=1}^T \gamma^\tau \sum_{k=1}^{2\times N_{lat} \times N_{lon}} \text{CRPS} \left( \hat{X_k}(t+\tau), x_k(t+\tau) \right) \\
    \label{eq:crps-loss}
\end{align}

To address the greater loss values at longer forecast leads, we introduce a power-law decaying weight over lead time, $\gamma^\tau$, where $\gamma=0.65$ is an empirically set hyper-parameter.

In addition, we use the AdamW adaptive gradient algorithm \citep{loshchilov2019} in conjunction with cosine annealing \citep{loshchilov2017} to dynamically adjust the learning rate during the training phase.
% Batch size of 64 with lr=1e-4 - 3e-7

\subsection{Evaluation metrics} \label{sec:evaluation_metrics}

Our analysis of the models, all of which generate ensemble member predictions, is based on probabilistic metrics as well as deterministic metrics of their ensemble mean. We evaluate all models on the test set (200 years) using SSTA and SSHA in the tropical Pacific. 

\paragraph{Anomaly correlation coefficient (ACC)}
The ACC is the temporal correlation between the model forecast $\hat{x}$ and the target $x$ at each spatial location $k$ over the test set. The ACC is defined as,

\begin{equation}
    \text{ACC}_k = \frac{\text{cov}(\hat{x}_k, x_k)}{\hat{\sigma}_{k, \tau} \sigma_k}
\end{equation}

where $\text{cov}(x_k(\tau), x_k) = \langle \hat{x}_k(tau), x \rangle$ is the covariance between the time series and $\sigma_k$ their respective variance.

\paragraph{Root mean square error skill score (RMSESS)} 
The RMSESS is a deterministic metric that compares the RMSE of the model to the RMSE of a reference forecast. Throughout this work, we choose the climatology as our reference forecast. The RMSESS can be written as

\begin{equation}
    \text{RMSESS}(\tau) = 1 - \frac{\text{RMSE}_\text{model} (\tau)}{\text{RMSE}_\text{ref} (\tau)} 
    = 1 - \frac{\sqrt{\frac{1}{N} \sum_{t=1}^N \left( \hat{\mathbf{x}}(t+\tau) - \mathbf{x}(t+\tau) \right)^2}}{\sigma_\mathbf{x}}
\end{equation}

where $\sigma_\mathbf{x}$ is the variance of the data. An RMSESS of 1 is a perfect model forecast and 0 is as good as the climatology forecast.

\paragraph{Continuous ranked probability skill score (CRPSS)}

Equivalently to the skill score of the RMSE, we define the CRPS skill score using Eq. \ref{eq:crps} as, $\text{CRPSS}(\hat{X}_k, x_k) = 1 - {\text{CRPS}_\text{model}(\hat{X}_k, x_k)} / {\text{CRPS}_\text{ref}(\hat{X}_{ref,k}, x_k)}$, where the reference forecast $\hat{X}_{ref,k}$ is the climatology. A CRPSS of 1 is a perfect model forecast and 0 is as good as the climatology forecast.

%%%%%%%%%%%%%%%%%%%%%%%%%%%%%%%%%%%%%%%%%%%%%%%%%%%%%%%%%%%%%%%%%%%%%
% ACKNOWLEDGMENTS
%%%%%%%%%%%%%%%%%%%%%%%%%%%%%%%%%%%%%%%%%%%%%%%%%%%%%%%%%%%%%%%%%%%%%
\newpage
\section*{Acknowledgements}
The authors acknowledge funding by the Deutsche Forschungsgemeinschaft (DFG, German Research Foundation) under Germany's Excellence Strategy -- EXC number 2064/1 -- Project number 390727645. Furthermore, we express our gratitude to the NOAA Physical Science Laboratory for making their resources available for this study. Antonietta Capotondi was supported by the NOAA Climate Program Office MAPP and CVP programs, and by  DOE Award DE-SC0023228. We further thank the International Max Planck Research School for Intelligent Systems for supporting Jakob Schlör.

%%%%%%%%%%%%%%%%%%%%%%%%%%%%%%%%%%%%%%%%%%%%%%%%%%%%%%%%%%%%%%%%%%%%%
% DATA AVAILABILITY STATEMENT
%%%%%%%%%%%%%%%%%%%%%%%%%%%%%%%%%%%%%%%%%%%%%%%%%%%%%%%%%%%%%%%%%%%%%
\section*{Data Availability Statement}

All data used in this study are publicly available and are referenced in the main text or the supplementary materials. Our code is available at \url{https://github.com/jakob-schloer/hybridLIM}.

%%%%%%%%%%%%%%%%%%%%%%%%%%%%%%%%%%%%%%%%%%%%%%%%%%%%%%%%%%%%%%%%%%%%%
% REFERENCES
%%%%%%%%%%%%%%%%%%%%%%%%%%%%%%%%%%%%%%%%%%%%%%%%%%%%%%%%%%%%%%%%%%%%%
\clearpage
\bibliography{references}

\clearpage
\appendix

% Change figure labelling
\renewcommand\thefigure{S\arabic{figure}}  
\setcounter{figure}{0} 

% Change table labelling
\renewcommand\thetable{S\arabic{figure}}  
\setcounter{table}{1} 

\section{Supplementary Materials} \label{SI-sec:SI}

\begin{figure}[!htp]
    \centering
    \includegraphics[width=.9\textwidth]{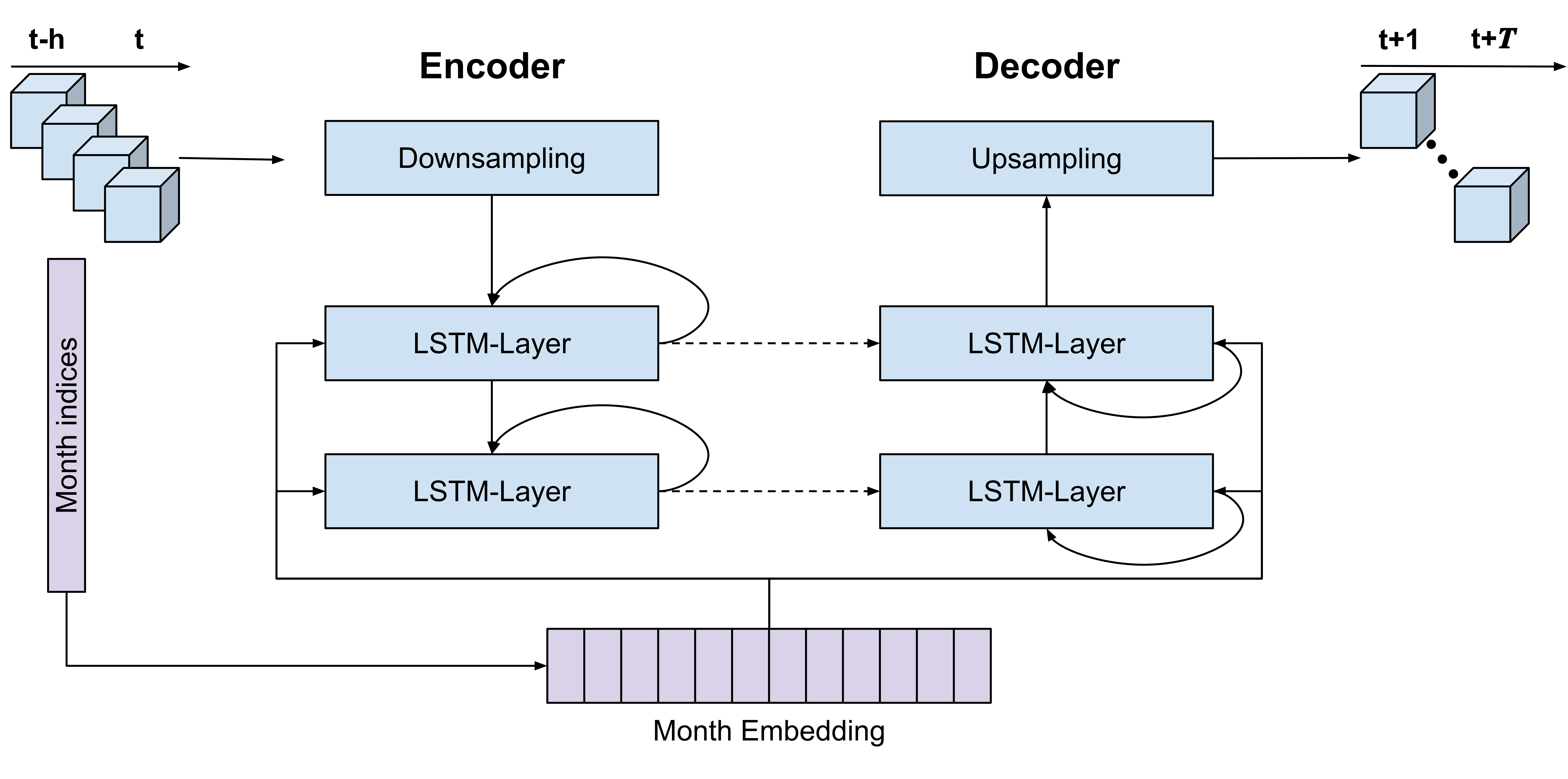}
    \caption{\textbf{Encoder-Decoder architecture of the LSTM and ConvLSTM.} The encoder network starts with a downsampling operation, which is represented by the EOF-truncation for the PC-LSTM and a strided convolution for the ConvLSTM. After downsampling, the input is linearly projected onto a higher-dimensional latent representation. This is sequentially processed through two LSTM (ConvLSTM) layers, integrating the transformed input with the preceding hidden state. The resulting hidden state at time t is passed to the decoder network. Reflecting the encoder's design, the decoder is composed of two LSTM (ConvLSTM) layers, extending over the future prediction span t + T, transferring the hidden state across time without further input. A final upsampling phase, using individual upsampling layers for each m-ensemble, members reverts the hidden state to the original input space. Monthly conditioning is incorporated into both encoder and decoder LSTM layers via affine transformations of the month embedding. }
    \label{SI-fig:architecture_lstm}
\end{figure}

\begin{figure}[!htp]
    \centering
    \includegraphics[width=1.0\textwidth]{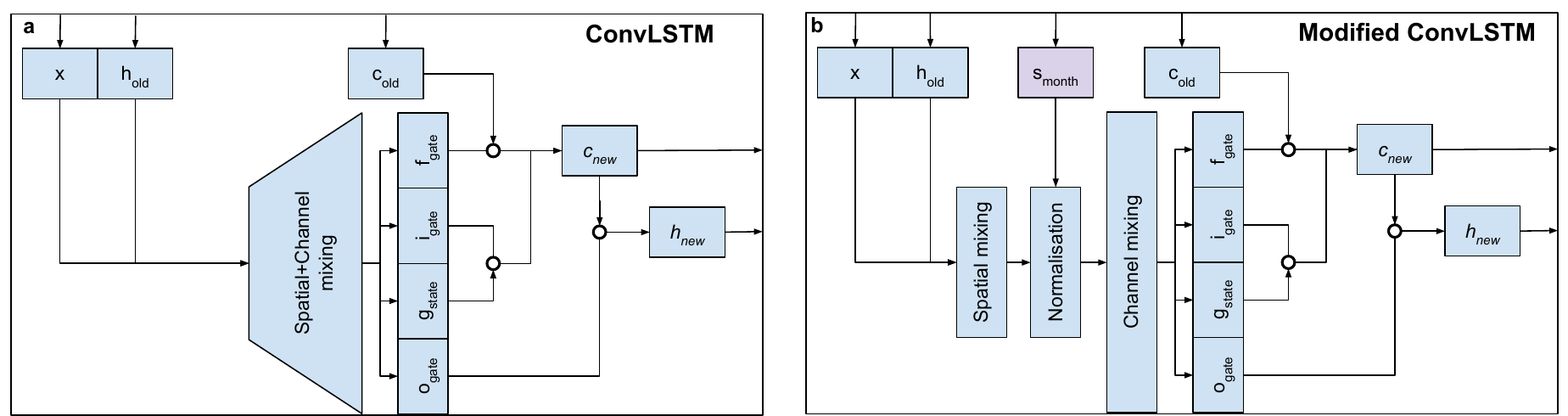}
    \caption{\textbf{Schematic representation of the ConvLSTM.} Input to the ConvLSTM cell (a) by \citet{shi2015} is the concatenated latent state $x$ and hidden state from the previous time step $h_\text{old}$. While in the ConvLSTM cell spatial and channel mixing is performed at once, we separate spatial and channel mixing into two convolutions (b), which drastically reduces the number of parameters. We further apply a layer normalization and conditioning on the monthly embedding (see Eq.~\ref{eq:film}) in between the spatial and channel mixing. The remainder of the cells are equivalent to the standard LSTM. }
    \label{SI-fig:architecture_swinlstm}
\end{figure}

\subsection{Seasonal skill dependency}
  \label{SI-sec:seasonal_skill}

\begin{figure}[!t]
    \centering
    \includegraphics[width=1.0\textwidth]{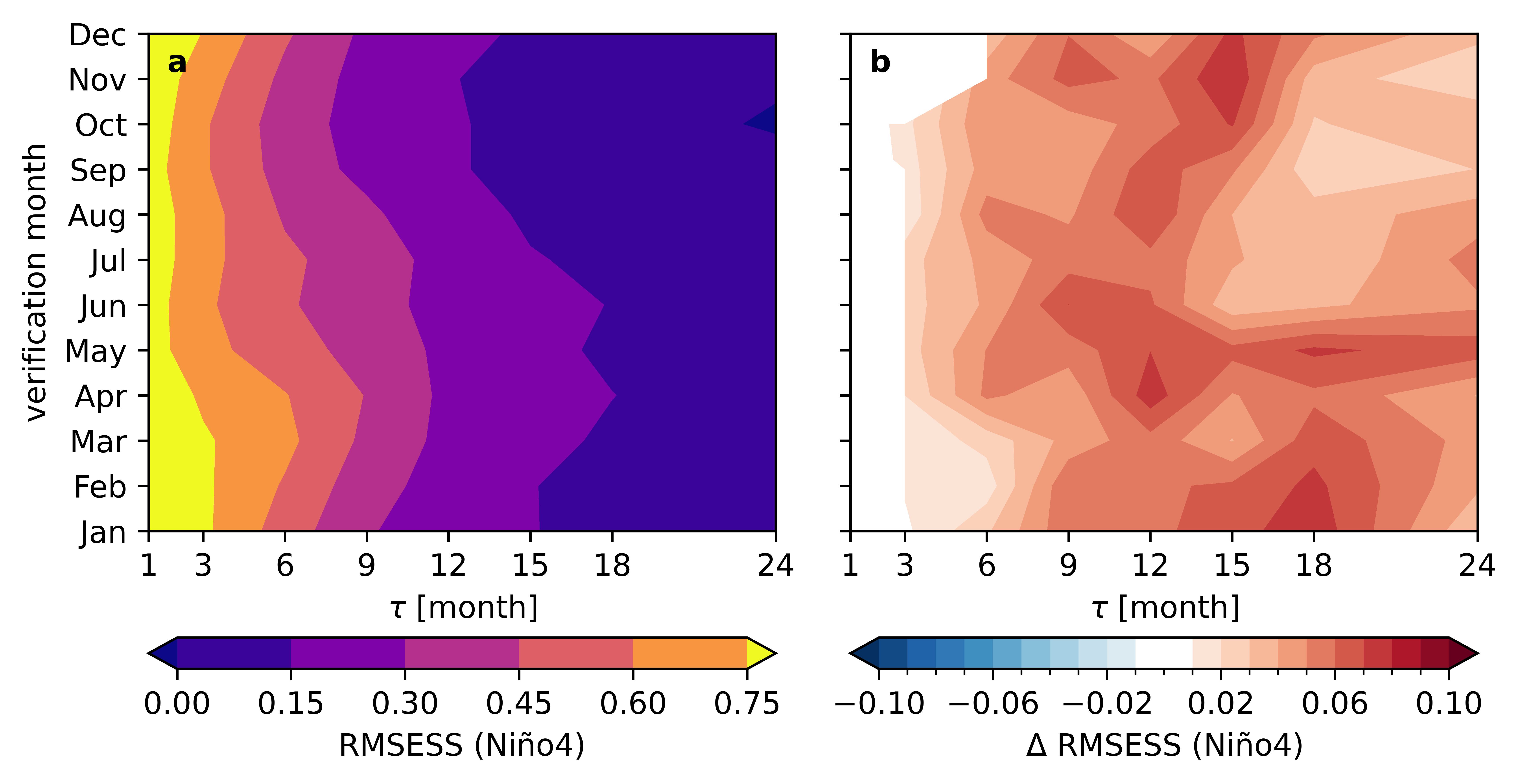}
    \caption{\textbf{Seasonal skill dependency of Niño4.} The average RMSE skill score of the CS-LIM forecast for Niño4 SSTA is analyzed over various lead times and verification months (a). The LIM-LSTM model forecast significantly improves relative to the CS-LIM, particularly for lead times between 9 and 18 months (b). In both panels, statistical significance is determined using a two-sided t-test on 1000 bootstrapped means of CS-LIM and LIM-LSTM model forecasts. Results displayed exceed the 95\% confidence interval. }
    \label{SI-fig:seasonal_skill}
\end{figure}

In addition to assessing the spatial distribution of skill (sec.~\ref{sec:skill_improvement}), we investigate the seasonal skill variation. The average RMSESS of the Niño4 SSTA from the CS-LIM forecast, evaluated over different lead times and verification months, indicates that late winter and spring months are better predicted than the late summer and fall (Fig.~\ref{SI-fig:seasonal_skill}a). These results are consistent with the findings by \citet{shin2021}, and align with the phenomenon known as the spring predictability barrier, characterized by a notable drop in the autocorrelation of the tropical Pacific SSTA in boreal spring.

For the LIM-LSTM model forecast, we observe the most significant RMSESS improvements upon the CS-LIM at lead times ranging between 9 and 18 months (Fig.~\ref{SI-fig:seasonal_skill}b). Specifically, the maximum enhancements are seen at 15 and 18 months during the winter months (December to February), while in spring, the peak skill improvement is at a lead time of 12 months.

%TC:endignore
\end{document}